%% file: emnlp2020.tex
\documentclass[11pt,a4paper]{article}
\usepackage[hyperref]{emnlp2020}
\usepackage{times}
\usepackage{latexsym}
\usepackage{url}
\usepackage{multirow}
\usepackage{graphicx}
\usepackage{booktabs}
\usepackage{float}
\usepackage{amsmath}
\usepackage{amssymb}
\usepackage{arydshln}
\graphicspath{ {.} }

\usepackage{microtype}

\aclfinalcopy 
\def\aclpaperid{3085} 

\title{No Answer is Better Than Wrong Answer: A Reflection Model for Document Level Machine Reading Comprehension}

\author{
{Xuguang Wang}$^1$ \quad {Linjun Shou}$^1$ \quad {Ming Gong}$^1$ \quad {Nan Duan}$^2$ \quad {Daxin Jiang}$^1$\footnotemark[3] \\
$^1${Microsoft STCA NLP Group, Beijing, China} \\
$^2${Microsoft Research Asia, Beijing, China} \\
\texttt{\{xugwang,lisho,migon,nanduan,djiang\}@microsoft.com}}

\date{}

\begin{document}
\maketitle
\footnotetext[3]{Corresponding author.} 
\begin{abstract}
The Natural Questions (NQ) benchmark set brings new challenges to Machine Reading Comprehension: the answers are not only at different levels of granularity (long and short), but also of richer types (including no-answer, yes/no, single-span and multi-span). In this paper, we target at this challenge and handle all answer types systematically. In particular, we propose a novel approach called Reflection Net which leverages a two-step training procedure to identify the no-answer and wrong-answer cases.
Extensive experiments are conducted to verify the effectiveness of our approach. At the time of paper writing (May.~20,~2020), our approach achieved the top 1 on both long and short answer leaderboard\footnote{\url{https://ai.google.com/research/NaturalQuestions/leaderboard}}, with F1 scores of 77.2 and 64.1, respectively.
\end{abstract}

\section{Introduction}
\label{intro}

Deep neural network models, such as ~\cite{cui-etal-2017-attention,chen-etal-2017-reading,clark-gardner-2018-simple,wang-etal-2018-multi-granularity,devlin-etal-2019-bert,liu2019roberta,yang2019xlnet}, have greatly advanced the state-of-the-arts of machine reading comprehension (MRC). Natural Questions (NQ)~\cite{kwiatkowski-etal-2019-natural} is a new Question Answering benchmark released by Google, which brings new challenges to the MRC area. One challenge is that the answers are provided at two-level granularity, i.e., long answer (e.g., a paragraph in the document) and short answer (e.g., an entity or entities in a paragraph). Therefore, the task requires the models to search for answers at both document level and passage level. Moreover, there are richer answer types in the NQ task. In addition to indicating textual answer spans (long and short), the models need to handle cases including no-answer (51\%), multi-span short answer (3.5\%), and yes/no (1\%) answer. Table \ref{table:nq_example} shows several examples in NQ challenge.
\input{data_example_tab}
\input{model_pic.tex}

Several works have been proposed to address the challenge of providing both long and short answers. \newcite{kwiatkowski-etal-2019-natural} adopts a pipeline approach, where a long answer is first identified from the document and then a short answer is extracted from the long answer. Although this approach is reasonable, it may lose the inherent correlation between the long and the short answer, since they are modeled separately. Several other works propose to model the context of the whole document and jointly train the long and short answers. For example, \newcite{alberti2019bert} split a document into multiple training instances using sliding windows, and leverages the overlapped tokens between windows for context modeling. A MRC model based on BERT~\cite{devlin-etal-2019-bert}  is applied to model long and short answer span jointly.
While previous approaches have proved effective to improve the performance on the NQ task, few works focus on the challenge of rich answer types in this QA set. We note that 51\% of the questions have no answer in the NQ set, therefore, it is critical for the model to accurately predict when to output the answer. 
For other answer types, such as multi-span short answer or yes/no answer, although they have a small percentage in the NQ set, they should not be ignored. Instead, a systematic design which can handle all kinds of answer types well would be more preferred in practice.

In this paper, we target the challenge of rich answer types, and particularly for no-answer. In particular, we first train an all types handling MRC model. Then, we leverage the trained MRC model to inference all the training data, train a second model, called the {\em Reflection model} takes as inputs the predicted answer, its context and MRC head features to predict a more accurate confidence score which distinguish the right answer from the wrong ones.
There are three reasons of applying a second-phase Reflection model. Firstly, the common practice of MRC confidence computing is based on heuristics of logits, which isn't normalized and isn't very comparable between different questions.\cite{chen-etal-2017-reading,alberti2019bert} Secondly, when training long document MRC model, the negative instances are down sampled by a large magnitude because they are too many compared with positive ones (see Section \ref{ss:mrc_model}). But when predicting, MRC model should inference all the instances. This data distribution discrepancy of train and predict result in that MRC model may be puzzled by some negative instance and predict a wrong answer with a high confidence score. Thirdly, MRC model learns the representation towards the relation between the question, its type, and the answer which isn't aware of the correctness of the predicted answer. Our second-phase model addresses these three issues and is similar to a reflection process which become the source of its name.
To the best of our knowledge, this is the first work to model all answer types in NQ task.
We conducted extensive experiments to verify the effectiveness of our approach. Our model achieved top 1 performance on both long and short answer leaderboard of NQ Challenge at the time of paper writing (May.~20,~2020). The F1 scores of our model were 77.2 and 64.1, respectively, improving over the previous best result by 1.1 and 2.7.

\section{Our Approach}\label{sec:approach}

We propose {\em Reflection Net} (see Figure~\ref{fig:model}), which consists of a MRC model for answer prediction and a Reflection model for answer confidence.

\subsection{MRC Model}
\label{ss:mrc_model}
Our MRC model (see Figure \ref{fig:model}(b)) is based on pre-trained transformers \cite{devlin-etal-2019-bert,liu2019roberta,alberti2019bert}, and it is able to handle all answer types in NQ challenge. 
We adopt the sliding window approach to deal with long document~\cite{alberti2019bert}, which slices the whole document into overlapping sliding windows. We pair each window with the question to get one training instance limiting the length to 512. The instances divide into positive ones whose window contain the answer and negative ones whose window doesn't contain. Since the documents are usually very long, there are too many negative instances. For efficient training, we down-sample the negative instances to some extent.

The targets of our MRC model include answer type and answer spans, which are denoted as $l = (t, s, e, ms)$. $t$ is the answer type, which can be one of the answer types described before or the special ``no-answer''. $s$ and $e$ are the start and end positions of the minimum single span that contains the corresponding answer. All answer types in NQ have a minimum single span~\cite{alberti2019bert}.
When answer type is multi-span, $ms$ represents the sequence labels of this answer, otherwise null. We adopt the B, I, O scheme to indicate multi-span answer~\cite{li2016dataset} in which $ms = (n_1, \dots, n_T)$, where $n_i \in \{\mbox{B, I, O}\}$. 
Then, the architecture of our MRC model is illustrated as following.
The input instance  $\mathbf{x} = (x_1, \dots, x_T)$ of the MRC model has the embedding:
\begin{equation}
\operatorname{E}(\mathbf{x}) = (\operatorname{E}(x_1), \dots, \operatorname{E}(x_T)),
\end{equation}
where
\begin{align}
\operatorname{E}(x_i)=\operatorname{E}_w(x_i) + \operatorname{E}_p(x_i) + \operatorname{E}_s(x_i),  
\label{embedding_input}
\end{align}
and $\operatorname{E}_w$, $\operatorname{E}_p$ and $\operatorname{E}_s$ are the operation of word embedding, positional embedding and segment embedding, respectively. The contextual hidden representation of the input sequence is 
\begin{equation}
h(\mathbf{x}) = \operatorname{T}_\theta(\operatorname{E}(\mathbf{x})) = (h(x_1), \dots, h(x_T)) 
\end{equation}
where $\operatorname{T}_\theta$ is pretrained Transformer~\cite{vaswani2017attention,devlin-etal-2019-bert,liu2019roberta} with parameter $\theta$. Next, we describe three types of model outputs.

\paragraph{Answer Type:}
Same with the method in~\newcite{kwiatkowski-etal-2019-natural},  we classify the hidden representation of [cls] token, $h(x_1)$ to answer types:
\begin{equation}
    p_{\text{type}} = \operatorname{softmax}(h(x_1)\cdot W_o^T)
\label{ans_type_p}
\end{equation}
where, $p_{\text{type}} \in \mathbb{R}^K$ is answer type probability, $K$ is the number of answer types , $h(x_1) \in \mathbb{R}^H$, $H$ is the size of hidden vectors in Transformer,  $W_o \in \mathbb{R}^{K\times H}$ is the parameters need to be learned. The loss of answer type prediction is:
\begin{equation}
    \mathcal{L}_{\text{type}} = -\operatorname{log}p_{\text{type=t}}
\end{equation}
where $t$ is the ground truth answer type.

\paragraph{Single Span:}
As described above, all kinds of answers have a minimal single span.
We model this target as predicting the start and end positions independently. For the no-answer case, we set the positions pointing to the [cls] token as in \newcite{devlin-etal-2019-bert}.
\begin{equation}
\label{start_p}
    p_{\text{start=i}} = \frac{\exp(S\cdot h(x_i))}{\sum_j \exp(S\cdot h(x_j))}
\end{equation}
\begin{equation}
\label{end_p}
    p_{\text{end=i}} = \frac{\exp(E\cdot h(x_i))}{\sum_j \exp(E\cdot h(x_j))}
\end{equation}
where $S \in \mathbb{R}^H, E \in \mathbb{R}^H$ are parameters need to be learned. The single span loss is:
\begin{equation}
    \mathcal{L}_{\text{span}} = -(\operatorname{log}p_{\text{start=s}} + \operatorname{log}p_{\text{end=e}})
\end{equation}

\paragraph{Multi Spans:}
We formulate the multi-spans prediction as a sequence labeling problem. To make the loss comparable with that for answer type and single span
, we do not use the traditional CRF or other sequence labeling loss, instead, directly feed the hidden representation of each token to a linear transformation and then classify to B, I, O labels:
\begin{equation}
    p_{\text{label}_i} = \operatorname{softmax}(h(x_i)\cdot W_l^T)
\end{equation}
where, $p_{\text{label}_i} \in \mathbb{R}^3$ is the B, I, O label probabilities of the $i$-th token. $W_l \in \mathbb{R}^{3\times H}$ is the parameter matrix. The loss of multi spans is:
\begin{equation}
    \mathcal{L}_{\text{multi-span}} = -\frac{1}{T}\sum_{i=1}^T\log p_{\text{label}_i=n_i}
\end{equation}
Combining all above three losses together, the total MRC model loss is denoted as:
\begin{equation}
    \mathcal{L}_{\text{mrc}} =
    \mathcal{L}_{\text{type}} + \mathcal{L}_{\text{span}} + \mathcal{L}_{\text{multi-span}}
\end{equation}
For cases which do not have multi-span answer, we simply set $\mathcal{L}_{\text{multi-span}}$ as $0$.

\input{features_table.tex}

Besides of predicting answer, MRC model should also output a corresponding confidence score.
In practice, we use the following heuristic \cite{alberti2019bert} to represent the confidence score of the predicted span:
\begin{equation}
    score = S\cdot h(x_s) + E\cdot h(x_e) - S\cdot h(x_1) - E\cdot h(x_1)
\label{heuris1}
\end{equation}
where $x_s, x_e, x_1$ are the predicted start, end and [cls] tokens, respectively. $S$ and $E$ are the learned parameters in Eq.~\ref{start_p} and \ref{end_p}.

To be specific of the answer prediction and confidence score calculation: firstly, we use MRC model to predict spans for all the sliding window instances of a document; then we rank predicted single spans based on its score Eq.~\eqref{heuris1}, choose the top 1 as predicted answer, and determine answer type based on probabilities of Eq.~\eqref{ans_type_p}, if the answer type is multi-span, we decode its corresponding sequence labels further; thirdly, we select as the long answer the DOM tree top level node containing the predicted top 1 span. The final confidence score of the predicted answer is its corresponding span score.

\subsection{Reflection Model}
\label{ss:aware_model}

Reflection model target a more precise confidence score which distinguish the right answer from two kinds of wrong ones (see Section~\ref{ss:ana}). The first one is predicting a wrong answer for a has-ans question, the second is predicting any answer for a no-ans question.

\paragraph{Training Data Generation:}
To generate Reflection model's training data, we leverage the trained MRC model above to inference its full training data (i.e. all the sliding window instances.): 

\begin{itemize}\setlength{\itemsep}{0pt} 
\item For all the instances belong to each one question, we \emph{only select the one} with top 1 predicted answer according to its confidence score.
\item The selected instance, MRC predicted answer, its corresponding head features described below and correctness label (if the predicted answer is same to the ground-truth answer, the label is 1; otherwise 0) together become a training case for Reflection model\footnote{When MRC model has predicted `no-answer', Reflection model throw away this question since the finial output is no-answer already determined.}. 
\end{itemize}

\paragraph{Model Training: }
As shown in Figure \ref{fig:model}(a), we initialize Reflection model with the parameters of the trained MRC model, and utilize a learning rate \emph{several times smaller} than the one used in MRC model. To directly receive important state information of the MRC model, we extract head features from the top layer of the MRC model when it is predicting the answer. As detailed in Table~\ref{tab:head}, score and ans\_type\_prob features are the two most straightforward ones; probabilities and logits features correspond to ``soft-targets" in knowledge distillation~\cite{hinton2015distilling}, which are so-called ``dark knowledge" with its distribution reflecting MRC model's inner state during answer prediction process. Here we only use top $n=5$ logits/probs instead of all. The head features are concatenated with the hidden representation of [cls] token, then followed by a hidden layer for final confidence prediction.

\paragraph{Formulation: }
Reflection model takes as inputs the selected instance $\mathbf{x}$ and the predicted answer. In detail, we create a dictionary $Ans$ whose elements are answer types and answer position marks\footnote{For NQ, it would be \{SINGLE\_SPAN, MULTI\_SPAN, YES, NO, LONG, START, END, B, I, O, EMPTY\}. }. We add answer type mark to the [cls] token, the position mark to corresponding tokens in position, and EMPTY to other tokens. The embedding representation of $i$-th token is given by:
\begin{equation}
    \operatorname{E}^r(x_i) = \operatorname{E}(x_i) + \operatorname{E}_r(f_i)
\end{equation}
where $r$ denotes Reflection model, $\operatorname{E}(x_i)$ is taken from Eq.~\eqref{embedding_input}, $f_i$ is one of $Ans$ element corresponding to token $x_i$ as described above, $\operatorname{E}_r$ is its embedding operation whose parameters is randomly initialized.
We use the same Transformer architecture as MRC model with parameter $\Phi$, denoted as $\operatorname{T}_\Phi$. The contextual hidden representations are given by:
\begin{equation}
    h^r(\mathbf{x}) = \operatorname{T_\Phi}(\operatorname{E}^r(\mathbf{x}))
\end{equation}

Then, we concatenate the [cls] token representation $h^r(x_1)$ with the head features, send it to a linear transformation activated with GELU~\cite{hendrycks2016gaussian} to get the aggregated representation as:
\begin{equation}
    \operatorname{hidden}(\mathbf{x}) = \operatorname{gelu}(\operatorname{concat}(h^r(x_1), \operatorname{head}(\mathbf{x}))\cdot W_r^T)
\label{aggregation}
\end{equation}
where, $W_r \in \mathbb{R}^{H\times (H+h)}$ is parameter matrix, $\operatorname{head}(\mathbf{x}) \in \mathbb{R}^{h}$ are head features\footnote{We transform head features by scale to $[0,1]$, sqrt, log, minus mean then divided by standard deviation.}. At last, we get the confidence score in probability:

\begin{equation}
    p_r = \operatorname{sigmoid}(A\cdot \operatorname{hidden}(\mathbf{x}))
\label{reflection_score}
\end{equation}
where $A \in \mathbb{R}^H$ is parameter vector. The loss is binary classification cross entropy given by:
\begin{equation}
    \mathcal{L}_{r} = -(y\cdot \log p_r + (1-y)\cdot \log(1-p_r))
\end{equation}
where, $y=1$ if MRC model's predicted answer (which is based on $\mathbf{x}$) is correct, otherwise 0. For inference, MRC model has to predict all sliding window instances of one document for each question, but Reflection model only needs to inference one instance who contains the MRC model predicted final answer. So the computation cost of Reflection model is very little.

\section{Experiments}\label{sec:exp}
\input{dev_nq_table.tex}
\input{leaderboard_table.tex}
We perform the experiments on NQ~\cite{kwiatkowski-etal-2019-natural} dataset which consists of 307,373 training examples, 7,830 development examples and 7,842 blind test examples used for leaderboard. The evaluation metrics are separated for long and short answers, each containing Precision (P), Recall (R), F-measure (F1) and Recall at fixed Precision (R@P=90, R@P=75, R@P=50). 
For each question, the system should provide both answer and its confidence score. During evaluation, the official evaluation script will calculate the optimal threshold which maximizes the F1, if answer score is higher than this threshold, the answer is triggered otherwise no-triggered.
Our dataset preprocessing method is similar to~\newcite{alberti2019bert}: firstly,  we tokenize the text according to different pretrained models, e.g. wordpiece for BERT,  BPE for RoBERTa; then use sliding window approach to slice document into instances as described in Section~\ref{ss:mrc_model}. For NQ, since the document is quite long, we add special atomic markup tokens to indicate which part of the document the model is reading.

\subsection{Implementation}
Our implementation is based on Huggingface Transformers ~\cite{wolf2019transformers}. All the pretrained models are large version (24 layers, 1024 hidden size, 16 heads, etc.). For MRC model training, we firstly finetune it on squad2.0 \cite{rajpurkar-etal-2018-know} data and then continue to finetune on NQ data. For Reflection model, we firstly leverage the MRC model to generate training data, and then finetune Reflection model which is initialized by MRC model parameters. 
We use one MRC model to deal with all answer types in NQ, but two Reflection models, one for long answer, the other for short. We manually tune the hyperparameters based on dev data F1 and submit best models to NQ organizer for leaderboard, list the best setting in Appendices~\ref{sec:appendix}. Experiments are performed on 4 NVIDIA Tesla P40 24GB cards, both MRC and Reflection model can be trained within 48 hours. Dev data inference can be finished within 1 hour. Adam~\cite{KingmaB14} is used for optimization. 

\subsection{Baselines}
\label{ss:baselines}
The first baseline is DocumentQA~\cite{clark-gardner-2018-simple} proposed to address the multi-paragraph reading comprehension task.
The second baseline is DecAtt + DocReader which is a pipeline approach~\cite{kwiatkowski-etal-2019-natural} and decompose full document reading comprehension task to firstly select long answer and then extract short answer. The third baseline BERT\textsubscript{joint} is proposed by~\newcite{alberti2019bert} which is similar to our MRC model but that it omits yes/no, multi-span short answer and it doesn't have a confidence prediction model like Reflection model. The rest two baselines include a single human annotator (Single-human) and an ensemble of human annotators (Super-annotator).
\subsection{Results}

The dev set results are shown in Table~\ref{tab:nq_dev_results}. Middle block are our results where subscript ``all type'' denotes that our MRC model is able to handle all answer types. Considering all the metrics, our BERT\textsubscript{all type} alone already surpass all the three baselines. For BERT based models, our BERT\textsubscript{all type} surpass BERT\textsubscript{joint} which ignores yes/no, multi-span answers by F1 scores of 4.8 and 1.8 point for long and short answers respectively. This shows the effectiveness of addressing all answer types in NQ.
Compared with BERT\textsubscript{all type} and RoBERTa\textsubscript{all type}, our Reflection model can further boost model performance significantly by providing more accurate answer confidence score. Take RoBERTa\textsubscript{all type} as an example, our Reflection model improves the F1 scores of long and short answers by 2.9 and 3.1 points respectively which outperform the single human annotator results on both long and short answers. 
For ensemble, we train 3 RoBERTa\textsubscript{all type} models with different random seed. When predicting, per each question we combine the same answers by summing its confidence scores and then select the final answer which has the highest confidence score. For ``+ Reflection", we leverage the same shared Reflection models to provide confidence scores for these three MRC models predicted answers and conduct the same ensemble strategy. We see that Reflection model can further boost MRC ensemble due to a more precise and consistent score.

\input{analysis_tab.tex}
Table~\ref{tab:leaderboard_nq} shows the leaderboard result on sequestered test data. At the time we are writing this paper, there are 40+ submissions for each long and short answer leaderboard. We list the aforementioned two baselines: DecAtt + DocReader and BERT\textsubscript{joint}, top 3 performance submissions and our ensemble (Ensemble (3) + Reflection). We achieved top 1 on both long and short leaderboard.
In real production scenarios, the most practical metric is recall at a fixed high precision like R@P=90. For example, in search engine scenarios, question answering system should provide answers with a guaranteed high precision bar. In terms of R@P=90, our method surpasses top submissions by a large margin, 12.8 and 6.8 points for long and short answer respectively. 
\subsection{Analysis}
\label{ss:ana}

NQ contains question which has a answer (has-ans) and question has no answer (no-ans). For has-ans questions, good performing model should predict right-ans as much as possible and wrong-ans as little as possible or replace wrong-ans with no-ans to increase precision. For no-ans questions, the best is to always predict no-ans because predict any answer equals to wrong-ans. As shown in Table~\ref{tab:analysis}, the no-ans questions are about half in NQ (3222 for long, 4374 for short in dev set) which is challenge. MRC model (RoBERTa\textsubscript{all type}) though powerful has predicted a lot of wrong-ans in each scenario. With our Reflection model to provide a more accurate confidence score which is leveraged to determine answer triggering, the prediction count of wrong-ans is decreased and no-ans increased saliently, thus lead to the  improvement of evaluation metrics. The overall trend agree well with our paper title ``No answer is better than wrong answer". However, as we can see, the no-answer \& wrong-ans identification problem is hard and far from being solved: Ideally, all the wrong-ans case should be assigned a low confidence score thus identified as no-ans, which requires more powerful confidence models.

\section{Ablation Study}
\label{sec:ablation}
\subsection{Ablation on Answer Types}
\label{ss:ablation_answer_type}
\input{ablat_ans_type.tex}

As described in Section~\ref{ss:mrc_model}, our MRC model can deal with all answer types. We perform experiments to verify the effectiveness of dealing with these answer types in short answer, based on the same RoBERTa large MRC model architecture. As shown in Table~\ref{tab:ablat_ans_type}, without dealing with multi-spans answers results in a 0.8 point F1 drop. And without dealing with the yes/no answer leads to a 1.4 point F1 drop. When we neither deal with multi-spans nor yes/no answer types, but only address single-span answer, we get a 56.0 F1 score which is 2.2 point less than our all types handling model: RoBERTa\textsubscript{all type}. Note that the ratios of multi-spans and yes/no answer types are only 3.5\% and 1\% respectively. Thus 2.2 points gain is quite decent considering the low coverage of these answer types.
\input{ablation_table.tex}

\subsection{Ablation and Variant of Reflection Model}
\label{ss:ab_refl}

For ablation/variation experiments on Reflection model, we use the same MRC model: RoBERTa\textsubscript{all type} to predict answer, which means they have exactly the same answer but different confidence score. The results are shown in Table~\ref{tab:ablation}.

\paragraph{Comparison with Verifier: }
To compare with verifier~\cite{tan2018know,hu2019read+}, we build an analogue one by taking following steps upon Reflection model: remove head features, keep predicted answer input and initialize transformer with original RoBERTa large parameters. This setting corresponds to a RoBERTa based verifier. The result is shown in `` w/o head features \& init.'' row, although there is a 1.2 and 0.7 point F1 boost of long and short answers respectively, it is less effective than our Reflection model. This demonstrates that head features and parameter initialization from MRC model are very important for Reflection model performing well.

\paragraph{Effect of Head Features: }
Head features are manually crafted features based on MRC model as described in Section~\ref{ss:aware_model}. We believe these features contain state information that can be leveraged to predict accurate confidence score. To justify our assumption, we feed head features alone to a feedforward neural network (FNN) with one hidden layer sized 200 and one output neuron which produces the confidence score. For training this FNN, we use the same pipeline and training target as our Reflection model. The results are shown in ``only head features'' row. Note that the vector size of original head features is only 42, it is interesting that only this small sized head features and simple FNN can beat MRC model's heuristic confidence score by a salient margin, 1.1 and 1.7 point F1 for long and short answer respectively.

\paragraph{Head features \& MRC [cls]: }
We experiment with reuse of MRC model's transformer, that say, the [cls] representation of Reflection model is replaced with MRC model's. For training, we use the same pipeline as standard Reflection model but without predicted answer as extra input. Another thing is that we freeze the parameters of MRC model but only train aggregation Eq. \eqref{aggregation} and confidence score layer Eq. \eqref{reflection_score}, because the training target are quite different from MRC model, further training will hurt the accuracy of answer prediction. This configuration save a lot memory and computation cost of prediction: all the data only need to pass through one Transformer. The results show it can improve most of the metrics. However, the [cls] representation in MRC model targets at answer types classification which include no-answer but not predicted wrong-ans, the performance isn't as good as Reflection model.

\section{Related Work}\label{sec:related}
\paragraph{Machine Reading Comprehension: }
Machine reading comprehension~\cite{hermann2015teaching,chen-etal-2017-reading,rajpurkar-etal-2016-squad,clark-gardner-2018-simple} is mostly based on the attention mechanism~\cite{BahdanauCB14,vaswani2017attention} that take as input $\langle$question, paragraph$\rangle$, compute an interactive representation of them and predict the start and end positions of the answer. When dealing with no-answer cases, popular method is to jointly model the answer position probability and no-answer probability by a shared softmax normalizer~\cite{kundu-ng-2018-nil,clark-gardner-2018-simple,devlin-etal-2019-bert}, or independently model the answerability as a binary classification problem~\cite{hu2019read+,yang2019xlnet,liu2019roberta}.
For long document processing, there are pipeline approaches of IR + Span Extraction~\cite{chen-etal-2017-reading}, DecAtt + DocReader~\cite{kwiatkowski-etal-2019-natural}, sliding window approach~\cite{alberti2019bert} and recently proposed long sequence handling Transformers~\cite{Kitaev2020Reformer:,guo-etal-2019-star,Beltagy2020LongformerTL}

\paragraph{Answer Verifier: }
Answer verifier~\cite{tan2018know,hu2019read+} is proposed to validate the legitimacy of the answer predicted by MRC model. First a MRC model is trained to predict the candidate answer. Then a verification model takes question, answer sentence as input and further verifies the validity of the answer. Our method extends ideas of this work, but there are some main differences. 
The primary one is that our model takes as inputs answer, context and MRC model's state where an answer is generated. Another difference is that our model is based on transformer and is initialized with MRC.

\section{Conclusion}\label{sec:con}

In this paper, we propose a systematic approach to handle rich answer types in MRC. In particular, we develop a Reflection Model to address the no-answer/wrong-answer cases. The key idea is to train a second phase model and predict the confidence score of a predicted answer based on its content, context and the state of MRC model. Experiments show that our approach achieves the state-of-the-art results on the NQ set. Measured by F1 and R@P=90, and on both long and short answer, our method surpasses the previous top systems with a large margin. Ablation studies also confirm the effectiveness of our approach.

\bibliographystyle{acl_natbib}
\bibliography{emnlp2020}
\newpage
\input{append}
\end{document}

%% file: data_example_tab.tex
\begin{table}[t]
    \tiny
    \renewcommand{\arraystretch}{1.5}
    \centering
    \begin{tabular}{p{0.2cm}lp{4.9cm}}
    \hline
    \textbf{(a)} & \textbf{Question}: &\emph{who made it to stage 3 in american ninja warrior season 9}\\
    &\textbf{Wikipedia Page}: &\emph{American Ninja Warrior (season 9)}\\
    &\textbf{Long Answer}: &\emph{Results: Joe Moravsky (3:34.34), Najee Richardson (3:39:71) and Sean Bryan finished to go into Stage 3.} \\ 
    &\textbf{Short Answer}: &\emph{Joe Moravsky, Najee Richardson, Sean Bryan} \\
    \hdashline[1pt/1pt]
    \textbf{(b)}& \textbf{Question}: &\emph{why does queen Elizabeth sign her name Elizabeth r}\\
    &\textbf{Wikipedia Page}: &\emph{Royal sign-manual}\\
    &\textbf{Long Answer}: &\emph{The royal sign-manual usually consists of the sovereign’s regnal name (without number, if otherwise used), followed by the letter R for Rex (King) or Regina (Queen). Thus, the signs-manual of both Elizabeth I and Elizabeth II read Elizabeth R ... 
    } \\ 
    &\textbf{Short Answer}: &\emph{NULL} \\
    \hdashline[1pt/1pt]
    \textbf{(c)} & \textbf{Question}: &\emph{is an end of terraced house semi detached}\\
    &\textbf{Wikipedia Page}: &\emph{Terraced house}\\
    &\textbf{Long Answer}: &\emph{In the 21st century, Montréal has continued to build row houses at a high rate, with 62\% of housing starts in the metropolitan area being apartment or row units.[10]Apartment complexes, high-rises, and semi-detached homes are less popular in Montréal when compared to large Canadian cities ...
    } \\ 
    &\textbf{Short Answer}: &\emph{YES} \\
    \hline
    \end{tabular}
    \caption{Example of NQ challenge, short answer cases: (a) Multi-span answer, (b) No-answer, (c) Yes/No.}
    \label{table:nq_example}
\end{table}

%% file: model_pic.tex
\begin{figure*}[t]
    \centering
    \includegraphics[scale=0.46, viewport=0 80 1228 540, clip=true]{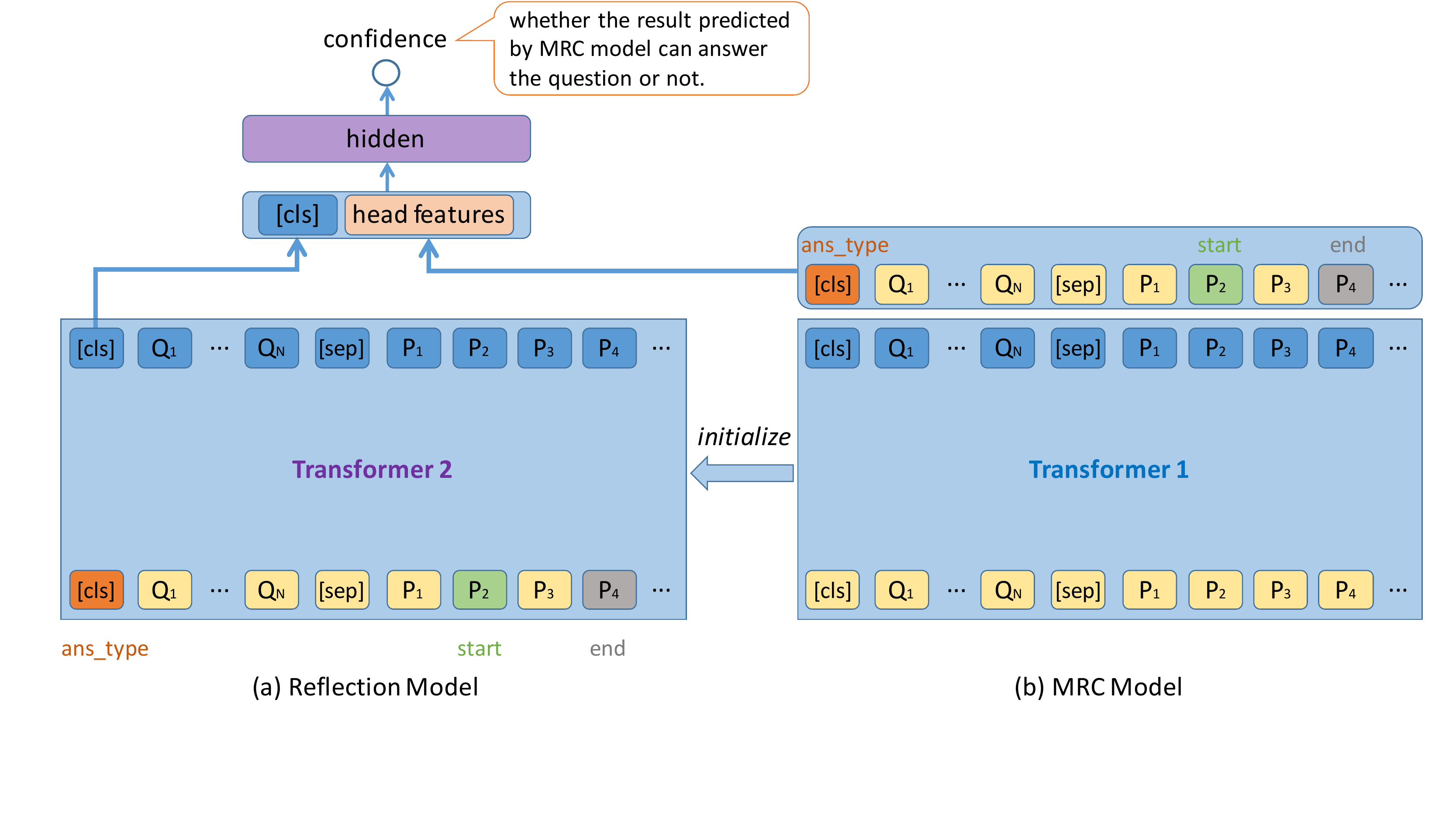}
    \caption{Overview of our proposed \emph{Reflection Net}, consisting of MRC model and its corresponding Reflection model. MRC model try its best to predict answer, Reflection model output corresponding answer confidence score.
    The left arrow denotes when training, Reflection model is initialized with the parameters of trained MRC model.
    }
    \label{fig:model}
\end{figure*}

%% file: features_table.tex
\begin{table*}[h!]
\begin{center}
\resizebox{\textwidth}{!}{%
 \begin{tabular}{|l|l|} 
 \hline
 \textbf{Feature name} & \multicolumn{1}{c|}{\textbf{Description}}  \\
 \hline
 \hline
 score & heuristic answer confidence score based on MRC model predictions, e.g. Eq.~\eqref{heuris1}   \\
\hline
 ans\_type & one-hot answer type feature. Answer type corresponding to the predicted answer is one, others are zeros. 
 \\
 \hline
 ans\_type\_probs & the probabilities of each answer type, e.g. Eq.~\eqref{ans_type_p} 
 \\
 \hline
 ans\_type\_prob & the probability of the answer type corresponding to the predicted answer.  \\
 \hline
 start\_logits & start logits of predicted answer, [cls] token and top $n$ start logits.  \\
 \hline
 end\_logits & end logits of predicted answer, [cls] token and top $n$ end logits. \\
 \hline
 start\_probs & start probabilities of predicted answer, [cls] token and top $n$ start probabilities.  \\
 \hline
 end\_probs & end probabilities of predicted answer, [cls] token and top $n$ end probabilities.  \\

  \hline
\end{tabular}}
\end{center}
\caption{Head Features: features extracted from the top layer of MRC model when it is on prediction mode. These features directly reflect some state information of MRC model's prediction process. }
\label{tab:head}
\end{table*}

%% file: dev_nq_table.tex
\begin{table*}[h!]
\begin{center}
\resizebox{\textwidth}{!}{%
 \begin{tabular}{l c c c c c c c c c c c c c} 
 \toprule
   & \multicolumn{6}{c}{NQ Long Answer Dev} &
   & \multicolumn{6}{c}{NQ Short Answer Dev} \\
   &    F1 &    P &   R &    R@P90 &    R@P75 &   R@P50 &
   &    F1 &    P &   R &    R@P90 &    R@P75 &   R@P50 \\
 \midrule
   DocumentQA
   & 46.1 & 47.5 & 44.7 & - & - & - &
   & 35.7 & 38.6 & 33.2 & - & - & - \\
   DecAtt + DocReader
   & 54.8 & 52.7 & 57.0 & - & - & - &
   & 31.4 & 34.3 & 28.9 & - & - & -  \\
   BERT\textsubscript{joint} 
   & 64.7 & 61.3 & 68.4 & - & - & - &
   & 52.7 & 59.5 & 47.3 & - & - & -  \\
 \midrule
   BERT\textsubscript{all type} 
   & 69.5 & 67.0 & 72.1 & 28.8 & 60.5 & 79.6 &
   & 54.5 & 60.6 & 49.5 & 0.0 & 33.1 & 54.8  \\
   \textbf{BERT\textsubscript{all type} + Reflection} 
   & \textbf{72.4} & \textbf{72.6} & \textbf{72.2} & \textbf{43.6} & \textbf{69.6} & \textbf{79.7} &
   & \textbf{56.1} & \textbf{64.3} & \textbf{49.7} & \textbf{14.3} & \textbf{40.3} & \textbf{56.4}  
   \\
   \hdashline
   RoBERTa\textsubscript{all type} 
   & 73.0 & 74.0 & 72.1 & 36.9 & 71.0 & 82.1 &
   & 58.2 & 63.3 & 53.9 & 19.0 & 42.6 & 61.2  \\
   \hspace{0.3cm}Ensemble (3) 
   & 73.6 & 71.8 & 75.4 & 37.3 & 71.6 & 83.5 &
   & 60.0 & 65.4 & 55.5 & 21.8 & 46.2 & 63.3  \\
   RoBERTa\textsubscript{all type}  + Reflection 
   & 75.9 & 79.4 & 72.7 & 52.7 & 75.5 & 82.1 &
   & 61.3 & 69.3 & 55.0 & 25.8 & 49.2 & 62.2   
   \\
   \hspace{0.3cm}\textbf{Ensemble (3) + Reflection} & \textbf{77.0} & \textbf{78.2} & \textbf{75.9} & \textbf{50.9} & \textbf{78.3} & \textbf{85.2} &
   & \textbf{63.4} & \textbf{67.9} & \textbf{59.4} & \textbf{29.0} & \textbf{52.9} & \textbf{66.2}
   \\
 \midrule
   Single-human
   & 73.4 & 80.4 & 67.6 & -    & -    & -    &
   & 57.5 & 63.4 & 52.6 & -    & -    & -    \\
   Super-annotator
   & 87.2 & 90.0 & 84.6 & -    & -    & -    &
   & 75.7 & 79.1 & 72.6 & -    & -    & -    \\
 \bottomrule
\end{tabular}}
\end{center}
\caption{NQ development set results. The top block rows are baselines we borrow from \newcite{alberti2019bert}. The last block rows are single human annotator and an ensemble of human annotators. The middle block are ours where BERT\textsubscript{all type} and RoBERTa\textsubscript{all type} are our MRC model. ``+ Reflection'' means that our Reflection model is used to provide answer confidence score. Ensemble (3) are three RoBERTa\textsubscript{all type} models.}
\label{tab:nq_dev_results}
\end{table*}

%% file: leaderboard_table.tex
\begin{table*}
\begin{center}
\resizebox{\textwidth}{!}{%
 \begin{tabular}{l c c c c c c c c c c c c c} 
 \toprule
   & \multicolumn{6}{c}{NQ Long Answer Test} &
   & \multicolumn{6}{c}{NQ Short Answer Test} \\
   &    F1 &    P &   R &    R@P90 &    R@P75 &   R@P50 &
   &    F1 &    P &   R &    R@P90 &    R@P75 &   R@P50 \\
 \midrule
   DecAtt + DocReader
   & 53.9 & 54.0 & 53.9 & 0.3 & 13.8 & 57.1 &
   & 29.0 & 32.7 & 26.1 & 0 & 0 & 0  \\
   BERT\textsubscript{joint} 
   & 66.2 & 64.1 & 68.3 & 22.6 & 47.2 & 76.6 &
   & 52.1 & 63.8 & 44.0 & 13.7 & 34.4 & 51.4  \\
 \midrule
   RoBERTa-mnlp-ensemble
   & 73.3 & 73.1 & 73.5 & 38.8 & 71.0 & 83.9 &
   & 61.4 & 69.6 & 54.9 & 28.2 & 50.4 & 62.7 \\
   RikiNet-ensemble
   & 75.6 & 75.3 & 75.9 & 40.5 & 76.0 & 85.2 &
   & 59.5 & 63.2 & 56.2 & 13.9 & 44.8 & 62.7\\
   RikiNet\_v2~\cite{liu-etal-2020-rikinet}
   & 76.1 & \textbf{78.1} & 74.2 & 40.1 & 77.0 & \textbf{85.7} &
   & 61.3 & 67.6 & 56.1 & 18.1 & 48.4 & 64.2\\
\midrule
   \textbf{ReflectionNet-ensemble}
   & \textbf{77.2} & 76.8 & \textbf{77.6} & \textbf{\underline{53.3}} & \textbf{78.5} & 85.2 &
   & \textbf{64.1} & \textbf{70.4} & \textbf{58.8} & \textbf{\underline{35.0}} & \textbf{54.4} & \textbf{66.1}\\
\bottomrule
\end{tabular}}
\end{center}
\caption{Leaderboard results (May.~20,~2020). The top block rows are baselines we described in Section~\ref{ss:baselines}. The middle rows are top 3 performance methods in leaderboard. The last is ours which achieved top 1 in both long and short answer leaderboard. Note that in terms of R@P=90 metric which is mostly used in real production scenarios, we surpass the top system by 12.8 and 6.8 absolute points for long and short answer respectively.}
\label{tab:leaderboard_nq}
\end{table*}

%% file: analysis_tab.tex
\begin{table*}[t]
\begin{center}
\resizebox{\textwidth}{!}{%
 \begin{tabular}{|l| c c c c |c c c| c c c c |c c|} 
 \hline
   & \multicolumn{6}{c}{NQ Long Answer Dev} &
   & \multicolumn{6}{c|}{NQ Short Answer Dev} \\
 \hline
 \hline
   Ground truth & \multicolumn{3}{c}{has-ans: 4608} && \multicolumn{2}{c}{no-ans: 3222} && \multicolumn{3}{c}{has-ans: 3456} && \multicolumn{2}{c|}{no-ans: 4374} \\
 \hline
   Model predict &  right-ans & wrong-ans & no-ans &&  wrong-ans &    no-ans &&  right-ans & wrong-ans & no-ans &&  wrong-ans &    no-ans \\
 \hline
 RoBERTa\textsubscript{all type} & 3324 & 446 & 838 && 725 & 2497 &&
 1863 & 561 & 1032 && 520 & 3854 \\
 \hline
 \multirow{2}{*}{\hspace{0.3cm} + Reflection} & 3347 & 334 & 927 && 534 & 2688 &&
 1908 & 441 & 1107 && 423 & 3951 \\
 & (+23) & (-112) & (+89) && (-191) & (+191) &&
 (+45) & (-120) & (+75) && (-97) & (+97) \\
 \hline
\end{tabular}}
\end{center}
\caption{The count of model predictions categorized as right-ans, wrong-ans and no-ans. Compared with RoBERTa\textsubscript{all type}, Reflection model leads to the decrease of wrong-ans and increase of no-ans and right-ans.}
\label{tab:analysis}
\end{table*}

%% file: ablat_ans_type.tex
\begin{table}[h]
\centering
\resizebox{0.47\textwidth}{!}{%
\begin{tabular}{l c c c c c} 
 \toprule
   & \multicolumn{5}{c}{NQ Short Answer Dev}  \\
   &    F1 &    P &   R &    R@P90 &   R@P50 \\
 \midrule
   RoBERTa\textsubscript{all type} 
   & 58.2 & 63.3 & 53.9 & 19.0 & 61.2  \\
 \midrule
   \hspace{0.1cm} - multi-spans (3.5\%)
   & 57.4 & 61.2 & 54.1 & 17.3  & 60.7 \\
   \hspace{0.1cm} - yes/no (1\%)
   & 56.8 & 62.8 & 51.9 & 17.1 & 58.5 \\
   \hspace{0.1cm} - multi-spans \& yes/no
   & 56.0 & 63.0 & 50.4 & 15.7 & 58.2 \\

 \bottomrule
\end{tabular}}
\caption{Ablation study on answer types. We compare all answer types handling model with ablation of multi-spans, yes/no type and both.}
\label{tab:ablat_ans_type}
\end{table}

%% file: ablation_table.tex
\begin{table*}[t]
\begin{center}
\resizebox{\textwidth}{!}{%
 \begin{tabular}{l c c c c c c c c c c c c c} 
 \toprule
   & \multicolumn{6}{c}{NQ Long Answer Dev} &
   & \multicolumn{6}{c}{NQ Short Answer Dev} \\
   &    F1 &    P &   R &    R@P90 &    R@P75 &   R@P50 &
   &    F1 &    P &   R &    R@P90 &    R@P75 &   R@P50 \\
 \midrule
   RoBERTa\textsubscript{all type} 
   & 73.0 & 74.0 & 72.1 & 36.9 & 71.0 & 82.1 &
   & 58.2 & 63.3 & 53.9 & 19.0 & 42.6 & 61.2  \\
   \textbf{RoBERTa\textsubscript{all type} + Reflection} 
   & 75.9 & 79.4 & 72.7 & 52.7 & 75.5 & 82.1 &
   & 61.3 & 69.3 & 55.0 & 25.8 & 49.2 & 62.2  \\
 \midrule
   \hspace{0.3cm} w/o~~head features \& init.
   & 74.2 & 76.7 & 71.9 & 45.5 & 73.1 & 82.0 &
   & 58.9 & 64.9 & 53.9 & 20.7 & 44.1 & 61.0  \\
   \hspace{0.3cm} only head features 
   & 74.1 & 74.3 & 73.9 & 39.0 & 72.8 & 82.1 &
   & 59.9 & 66.2 & 54.7 & 19.1 & 45.4 & 61.8  \\
   \hspace{0.3cm} head features \& MRC [cls]
   & 74.5 & 76.4 & 72.7 & 44.8 & 73.8 & 82.1 &
   & 60.1 & 64.2 & 56.5 & 21.6 & 45.8 & 61.9  \\

 \bottomrule
\end{tabular}}
\end{center}

\caption{Ablation and Variant of Reflection model. There are absence of head features and initialized from MRC model, simple three layer feedforward neural networks which take as input only head features, and lastly, head features integrated with MRC [cls] hidden representation.}
\label{tab:ablation}

\end{table*}

%% file: append.tex
\appendix
\section{Appendices: Hyperparameters}
\label{sec:appendix}
\begin{table}[htbp]
\resizebox{0.5\textwidth}{!}{%
 \begin{tabular}{l c c c c c } 
 \toprule
   \multirow{2}{*}{\textbf{Hyperparams}} & \multicolumn{2}{c}{BERT} && \multicolumn{2}{c}{RoBERTa} \\
   \cline{2-3}
   \cline{5-6}
    & \textbf{MRC} & \textbf{Reflection} && \textbf{MRC} & \textbf{Reflection}  \\
 \midrule
   Dropout & \multicolumn{2}{c}{0.1} && \multicolumn{2}{c}{0.1}\\
   Attention dropout & \multicolumn{2}{c}{0.1} && \multicolumn{2}{c}{0.1}\\
   Max sequence length & \multicolumn{2}{c}{512} && \multicolumn{2}{c}{512}\\
   Learning Rate &3e-5&5e-6&&2.2e-5&5e-6 \\ 
   Batch Size & \multicolumn{2}{c}{24} && \multicolumn{2}{c}{48}\\
   Weight Decay & \multicolumn{2}{c}{0.01} && \multicolumn{2}{c}{0.01}\\
   Epochs &1&2&&1&2\\
   Learning Rate Decay & \multicolumn{2}{c}{Linear} && \multicolumn{2}{c}{Linear}\\
   Warmup ratio & \multicolumn{2}{c}{0.1} && \multicolumn{2}{c}{0.06}\\
   Gradient Clipping & \multicolumn{2}{c}{1.0} && \multicolumn{2}{c}{-}\\
   Adam $\epsilon$ & \multicolumn{2}{c}{1e-8} && \multicolumn{2}{c}{1e-8}\\
   Adam $\beta_1$ & \multicolumn{2}{c}{0.9} && \multicolumn{2}{c}{0.9}\\
   Adam $\beta_2$ & \multicolumn{2}{c}{0.999} && \multicolumn{2}{c}{0.999}\\
 \bottomrule
\end{tabular}}
\caption{Training hyperparameters of MRC model and Reflection model.}
\label{tab:hyperparameters}
\end{table}

%% file: emnlp2020.bbl
\begin{thebibliography}{26}
\expandafter\ifx\csname natexlab\endcsname\relax\def\natexlab#1{#1}\fi

\bibitem[{Alberti et~al.(2019)Alberti, Lee, and Collins}]{alberti2019bert}
Chris Alberti, Kenton Lee, and Michael Collins. 2019.
\newblock A bert baseline for the natural questions.
\newblock \emph{arXiv preprint arXiv:1901.08634}.

\bibitem[{Bahdanau et~al.(2015)Bahdanau, Cho, and Bengio}]{BahdanauCB14}
Dzmitry Bahdanau, Kyunghyun Cho, and Yoshua Bengio. 2015.
\newblock \href {http://arxiv.org/abs/1409.0473} {Neural machine translation by
  jointly learning to align and translate}.
\newblock In \emph{3rd International Conference on Learning Representations,
  {ICLR} 2015, San Diego, CA, USA, May 7-9, 2015, Conference Track
  Proceedings}.

\bibitem[{Beltagy et~al.(2020)Beltagy, Peters, and
  Cohan}]{Beltagy2020LongformerTL}
Iz~Beltagy, Matthew~E. Peters, and Arman Cohan. 2020.
\newblock Longformer: The long-document transformer.
\newblock \emph{ArXiv}, abs/2004.05150.

\bibitem[{Chen et~al.(2017)Chen, Fisch, Weston, and
  Bordes}]{chen-etal-2017-reading}
Danqi Chen, Adam Fisch, Jason Weston, and Antoine Bordes. 2017.
\newblock \href {https://doi.org/10.18653/v1/P17-1171} {Reading {W}ikipedia to
  answer open-domain questions}.
\newblock In \emph{Proceedings of the 55th Annual Meeting of the Association
  for Computational Linguistics (Volume 1: Long Papers)}, pages 1870--1879,
  Vancouver, Canada. Association for Computational Linguistics.

\bibitem[{Clark and Gardner(2018)}]{clark-gardner-2018-simple}
Christopher Clark and Matt Gardner. 2018.
\newblock \href {https://doi.org/10.18653/v1/P18-1078} {Simple and effective
  multi-paragraph reading comprehension}.
\newblock In \emph{Proceedings of the 56th Annual Meeting of the Association
  for Computational Linguistics (Volume 1: Long Papers)}, pages 845--855,
  Melbourne, Australia. Association for Computational Linguistics.

\bibitem[{Cui et~al.(2017)Cui, Chen, Wei, Wang, Liu, and
  Hu}]{cui-etal-2017-attention}
Yiming Cui, Zhipeng Chen, Si~Wei, Shijin Wang, Ting Liu, and Guoping Hu. 2017.
\newblock \href {https://doi.org/10.18653/v1/P17-1055}
  {Attention-over-attention neural networks for reading comprehension}.
\newblock In \emph{Proceedings of the 55th Annual Meeting of the Association
  for Computational Linguistics (Volume 1: Long Papers)}, pages 593--602,
  Vancouver, Canada. Association for Computational Linguistics.

\bibitem[{Devlin et~al.(2019)Devlin, Chang, Lee, and
  Toutanova}]{devlin-etal-2019-bert}
Jacob Devlin, Ming-Wei Chang, Kenton Lee, and Kristina Toutanova. 2019.
\newblock \href {https://doi.org/10.18653/v1/N19-1423} {{BERT}: Pre-training of
  deep bidirectional transformers for language understanding}.
\newblock In \emph{Proceedings of the 2019 Conference of the North {A}merican
  Chapter of the Association for Computational Linguistics: Human Language
  Technologies, Volume 1 (Long and Short Papers)}, pages 4171--4186,
  Minneapolis, Minnesota. Association for Computational Linguistics.

\bibitem[{Guo et~al.(2019)Guo, Qiu, Liu, Shao, Xue, and
  Zhang}]{guo-etal-2019-star}
Qipeng Guo, Xipeng Qiu, Pengfei Liu, Yunfan Shao, Xiangyang Xue, and Zheng
  Zhang. 2019.
\newblock \href {https://doi.org/10.18653/v1/N19-1133} {Star-transformer}.
\newblock In \emph{Proceedings of the 2019 Conference of the North {A}merican
  Chapter of the Association for Computational Linguistics: Human Language
  Technologies, Volume 1 (Long and Short Papers)}, pages 1315--1325,
  Minneapolis, Minnesota. Association for Computational Linguistics.

\bibitem[{Hendrycks and Gimpel(2016)}]{hendrycks2016gaussian}
Dan Hendrycks and Kevin Gimpel. 2016.
\newblock Gaussian error linear units (gelus).
\newblock \emph{arXiv preprint arXiv:1606.08415}.

\bibitem[{Hermann et~al.(2015)Hermann, Kocisky, Grefenstette, Espeholt, Kay,
  Suleyman, and Blunsom}]{hermann2015teaching}
Karl~Moritz Hermann, Tomas Kocisky, Edward Grefenstette, Lasse Espeholt, Will
  Kay, Mustafa Suleyman, and Phil Blunsom. 2015.
\newblock Teaching machines to read and comprehend.
\newblock In \emph{Advances in neural information processing systems}, pages
  1693--1701.

\bibitem[{Hinton et~al.(2015)Hinton, Vinyals, and Dean}]{hinton2015distilling}
Geoffrey Hinton, Oriol Vinyals, and Jeff Dean. 2015.
\newblock Distilling the knowledge in a neural network.
\newblock \emph{arXiv preprint arXiv:1503.02531}.

\bibitem[{Hu et~al.(2019)Hu, Wei, Peng, Huang, Yang, and Li}]{hu2019read+}
Minghao Hu, Furu Wei, Yuxing Peng, Zhen Huang, Nan Yang, and Dongsheng Li.
  2019.
\newblock Read+ verify: Machine reading comprehension with unanswerable
  questions.
\newblock In \emph{Proceedings of the AAAI Conference on Artificial
  Intelligence}, volume~33, pages 6529--6537.

\bibitem[{Kingma and Ba(2015)}]{KingmaB14}
Diederik~P. Kingma and Jimmy Ba. 2015.
\newblock \href {http://arxiv.org/abs/1412.6980} {Adam: {A} method for
  stochastic optimization}.
\newblock In \emph{3rd International Conference on Learning Representations,
  {ICLR} 2015, San Diego, CA, USA, May 7-9, 2015, Conference Track
  Proceedings}.

\bibitem[{Kitaev et~al.(2020)Kitaev, Kaiser, and
  Levskaya}]{Kitaev2020Reformer:}
Nikita Kitaev, Lukasz Kaiser, and Anselm Levskaya. 2020.
\newblock \href {https://openreview.net/forum?id=rkgNKkHtvB} {Reformer: The
  efficient transformer}.
\newblock In \emph{International Conference on Learning Representations}.

\bibitem[{Kundu and Ng(2018)}]{kundu-ng-2018-nil}
Souvik Kundu and Hwee~Tou Ng. 2018.
\newblock \href {https://doi.org/10.18653/v1/D18-1456} {A nil-aware answer
  extraction framework for question answering}.
\newblock In \emph{Proceedings of the 2018 Conference on Empirical Methods in
  Natural Language Processing}, pages 4243--4252, Brussels, Belgium.
  Association for Computational Linguistics.

\bibitem[{Kwiatkowski et~al.(2019)Kwiatkowski, Palomaki, Redfield, Collins,
  Parikh, Alberti, Epstein, Polosukhin, Devlin, Lee, Toutanova, Jones, Kelcey,
  Chang, Dai, Uszkoreit, Le, and Petrov}]{kwiatkowski-etal-2019-natural}
Tom Kwiatkowski, Jennimaria Palomaki, Olivia Redfield, Michael Collins, Ankur
  Parikh, Chris Alberti, Danielle Epstein, Illia Polosukhin, Jacob Devlin,
  Kenton Lee, Kristina Toutanova, Llion Jones, Matthew Kelcey, Ming-Wei Chang,
  Andrew~M. Dai, Jakob Uszkoreit, Quoc Le, and Slav Petrov. 2019.
\newblock \href {https://doi.org/10.1162/tacl_a_00276} {Natural questions: A
  benchmark for question answering research}.
\newblock \emph{Transactions of the Association for Computational Linguistics},
  7:453--466.

\bibitem[{Li et~al.(2016)Li, Li, He, Wang, Cao, Zhou, and Xu}]{li2016dataset}
Peng Li, Wei Li, Zhengyan He, Xuguang Wang, Ying Cao, Jie Zhou, and Wei Xu.
  2016.
\newblock Dataset and neural recurrent sequence labeling model for open-domain
  factoid question answering.
\newblock \emph{arXiv preprint arXiv:1607.06275}.

\bibitem[{Liu et~al.(2020)Liu, Gong, Fu, Yan, Chen, Jiang, Lv, and
  Duan}]{liu-etal-2020-rikinet}
Dayiheng Liu, Yeyun Gong, Jie Fu, Yu~Yan, Jiusheng Chen, Daxin Jiang, Jiancheng
  Lv, and Nan Duan. 2020.
\newblock \href {https://doi.org/10.18653/v1/2020.acl-main.604} {{R}iki{N}et:
  Reading {W}ikipedia pages for natural question answering}.
\newblock In \emph{Proceedings of the 58th Annual Meeting of the Association
  for Computational Linguistics}, pages 6762--6771, Online. Association for
  Computational Linguistics.

\bibitem[{Liu et~al.(2019)Liu, Ott, Goyal, Du, Joshi, Chen, Levy, Lewis,
  Zettlemoyer, and Stoyanov}]{liu2019roberta}
Yinhan Liu, Myle Ott, Naman Goyal, Jingfei Du, Mandar Joshi, Danqi Chen, Omer
  Levy, Mike Lewis, Luke Zettlemoyer, and Veselin Stoyanov. 2019.
\newblock Roberta: A robustly optimized bert pretraining approach.
\newblock \emph{arXiv preprint arXiv:1907.11692}.

\bibitem[{Rajpurkar et~al.(2018)Rajpurkar, Jia, and
  Liang}]{rajpurkar-etal-2018-know}
Pranav Rajpurkar, Robin Jia, and Percy Liang. 2018.
\newblock \href {https://doi.org/10.18653/v1/P18-2124} {Know what you don{'}t
  know: Unanswerable questions for {SQ}u{AD}}.
\newblock In \emph{Proceedings of the 56th Annual Meeting of the Association
  for Computational Linguistics (Volume 2: Short Papers)}, pages 784--789,
  Melbourne, Australia. Association for Computational Linguistics.

\bibitem[{Rajpurkar et~al.(2016)Rajpurkar, Zhang, Lopyrev, and
  Liang}]{rajpurkar-etal-2016-squad}
Pranav Rajpurkar, Jian Zhang, Konstantin Lopyrev, and Percy Liang. 2016.
\newblock \href {https://doi.org/10.18653/v1/D16-1264} {{SQ}u{AD}: 100,000+
  questions for machine comprehension of text}.
\newblock In \emph{Proceedings of the 2016 Conference on Empirical Methods in
  Natural Language Processing}, pages 2383--2392, Austin, Texas. Association
  for Computational Linguistics.

\bibitem[{Tan et~al.(2018)Tan, Wei, Zhou, Yang, Lv, and Zhou}]{tan2018know}
Chuanqi Tan, Furu Wei, Qingyu Zhou, Nan Yang, Weifeng Lv, and Ming Zhou. 2018.
\newblock I know there is no answer: Modeling answer validation for machine
  reading comprehension.
\newblock In \emph{CCF International Conference on Natural Language Processing
  and Chinese Computing}, pages 85--97. Springer.

\bibitem[{Vaswani et~al.(2017)Vaswani, Shazeer, Parmar, Uszkoreit, Jones,
  Gomez, Kaiser, and Polosukhin}]{vaswani2017attention}
Ashish Vaswani, Noam Shazeer, Niki Parmar, Jakob Uszkoreit, Llion Jones,
  Aidan~N Gomez, {\L}ukasz Kaiser, and Illia Polosukhin. 2017.
\newblock Attention is all you need.
\newblock In \emph{Advances in neural information processing systems}, pages
  5998--6008.

\bibitem[{Wang et~al.(2018)Wang, Yan, and
  Wu}]{wang-etal-2018-multi-granularity}
Wei Wang, Ming Yan, and Chen Wu. 2018.
\newblock \href {https://doi.org/10.18653/v1/P18-1158} {Multi-granularity
  hierarchical attention fusion networks for reading comprehension and question
  answering}.
\newblock In \emph{Proceedings of the 56th Annual Meeting of the Association
  for Computational Linguistics (Volume 1: Long Papers)}, pages 1705--1714,
  Melbourne, Australia. Association for Computational Linguistics.

\bibitem[{Wolf et~al.(2019)Wolf, Debut, Sanh, Chaumond, Delangue, Moi, Cistac,
  Rault, Louf, Funtowicz et~al.}]{wolf2019transformers}
Thomas Wolf, Lysandre Debut, Victor Sanh, Julien Chaumond, Clement Delangue,
  Anthony Moi, Pierric Cistac, Tim Rault, R{\'e}mi Louf, Morgan Funtowicz,
  et~al. 2019.
\newblock Transformers: State-of-the-art natural language processing.
\newblock \emph{arXiv preprint arXiv:1910.03771}.

\bibitem[{Yang et~al.(2019)Yang, Dai, Yang, Carbonell, Salakhutdinov, and
  Le}]{yang2019xlnet}
Zhilin Yang, Zihang Dai, Yiming Yang, Jaime Carbonell, Ruslan Salakhutdinov,
  and Quoc~V Le. 2019.
\newblock Xlnet: Generalized autoregressive pretraining for language
  understanding.
\newblock \emph{arXiv preprint arXiv:1906.08237}.

\end{thebibliography}
